\pdfoutput=1
\documentclass[11pt]{article}

\usepackage[]{acl2021}

\usepackage{times}
\usepackage{latexsym}

\usepackage[T1]{fontenc}
\usepackage[utf8]{inputenc}

\usepackage{microtype}

\usepackage{amsfonts}
\usepackage{amsmath}
\usepackage{graphicx}
\usepackage{booktabs}

\title{Code Synonyms Do Matter: \\ Multiple Synonyms Matching Network for Automatic ICD Coding}

\author{
Zheng Yuan$^{12}$\thanks{$\quad$Work done at Alibaba DAMO Academy.} \space\space\space
Chuanqi Tan$^{2}$ \space\space
Songfang Huang$^{2}$ \space\space\\
$^{1}$Tsinghua University \space\space\space\space
$^{2}$Alibaba Group\\
\texttt{yuanz17@mails.tsinghua.edu.cn}\\
\texttt{\{chuanqi.tcq,songfang.hsf\}@alibaba-inc.com}
}

\date{}

\begin{document}
\maketitle
\begin{abstract}
Automatic ICD coding is defined as assigning disease codes to electronic medical records (EMRs).
Existing methods usually apply label attention with code representations to match related text snippets.
Unlike these works that model the label with the code hierarchy or description, we argue that the code synonyms can provide more comprehensive knowledge based on the observation that the code expressions in EMRs vary from their descriptions in ICD. 
By aligning codes to concepts in UMLS, we collect synonyms of every code. Then, we propose a multiple synonyms matching network to leverage synonyms for better code representation learning, and finally help the code classification. 
Experiments on the MIMIC-III dataset show that our proposed method outperforms previous state-of-the-art methods.

% Therefore, we propose leveraging synonyms to better represent each code. 
% They use descriptions of codes to obtain codes representations.
% However, the code expressions in EMRs can be various and different from code descriptions from ICD.
% We argue that code synonyms in medical knowledge graphs can offer more knowledge of codes and provide possible matching examples to enhance the performances of automatic ICD coding.
% In this paper, we present a multiple synonyms matching network that utilizes code synonyms for better codes representations and label attention.
\end{abstract}

\section{Introduction}
International Classification of Diseases (ICD) is a classification and terminology that provides diagnostic codes with descriptions for diseases\footnote{\url{who.int/standards/classifications/classification-of-diseases}}.
The task of ICD coding refers to assigning ICD codes to electronic medical records (EMRs) which is highly related to clinical tasks or systems including patient similarity learning \cite{suo2018deep}, medical billing \cite{sonabend2020automated}, and clinical decision support systems \cite{sutton2020overview}. Traditionally, healthcare organizations have to employ specialized coders for this task, which is expensive, time-consuming, and error-prone. As a result, many methods have been proposed for automatic ICD coding since the 1990s \cite{de1998hierarchical}.

% Coders solve ICD coding via matching the relevant text snippets in EMRs with descriptions provided by ICD \cite{o2005measuring}.

% However, it is time-consuming for manual ICD coding
% Due to the expensive and time-consumed manual coding, automatic ICD coding has been extensively studied since the 1990s \cite{de1998hierarchical}.
% Inspired by coders, 

Recent methods treat this task as a multi-label classification problem \cite{xie2018neural,li2020icd,zhou2021automatic}, which learn deep representations of EMRs with an RNN or CNN encoder and predict codes with a multi-label classifier.
Recent state-of-the-art methods propose label attention that uses the code representations as attention queries to extract the code-related representations\footnote{``Label'' equals to ``code'' in some contexts of this paper.} \cite{mullenbach-etal-2018-explainable}.
Many works further propose using code hierarchical structures \cite{falis2019ontological,xie2019ehr,cao2020hypercore} and descriptions \cite{ijcai2020-556,kim2021read} for better label representations.

In this work, we argue that the synonyms of codes can provide more comprehensive information. For example, the description of code \textit{244.9} is ``Unspecified hypothyroidism'' in ICD. 
However, this code can be described in different forms in EMRs such as ``low t4'' and ``subthyroidism''. Fortunately, these different expressions can be found in the Unified Medical Language System \cite{Bodenreider2004}, a repository of biomedical vocabularies that contains various synonyms for all ICD codes. Therefore, we propose to leverage synonyms of codes to help the label representation learning and further benefit its matching to the EMR texts. 

To model the synonym and its matching to EMR text, we further propose a \textbf{M}ultiple \textbf{S}ynonyms \textbf{M}atching \textbf{N}etwork (\textbf{MSMN})\footnote{Our codes and model can be found at \url{https://github.com/GanjinZero/ICD-MSMN}.}.
Specifically, we first apply a shared LSTM to encode EMR texts and each synonym. Then, we propose a novel multi-synonyms attention mechanism inspired by the multi-head attention \cite{vaswani2017attention}, which considers synonyms as attention queries to extract different code-related text snippets for code-wise representations.
% For one specific code, we use different label descriptions from the code to query the EMRs representations to obtain the multi-head label attention representations.
Finally, we propose using a biaffine-based similarity of code-wise text representations and code representations for classification.

We conduct experiments on the MIMIC-III dataset with two settings: full codes and top-50 codes. Results show that our method performs better than previous state-of-the-art methods.
% MSMN performs better than previous state-of-the-art methods, achieving macro-AUC of 95.1 (+1.3) and precision@8 of 74.9 (+0.4)
% Additionally, we observe that the EMRs are usually too long. 
% One model can use random N-grams to overfit the dataset easily. 
% We agure that the regularization is necessary for ICD coding.
% We apply r-drop ...

\section{Approach}
\begin{figure}
\centering
\includegraphics[width=3.0in]{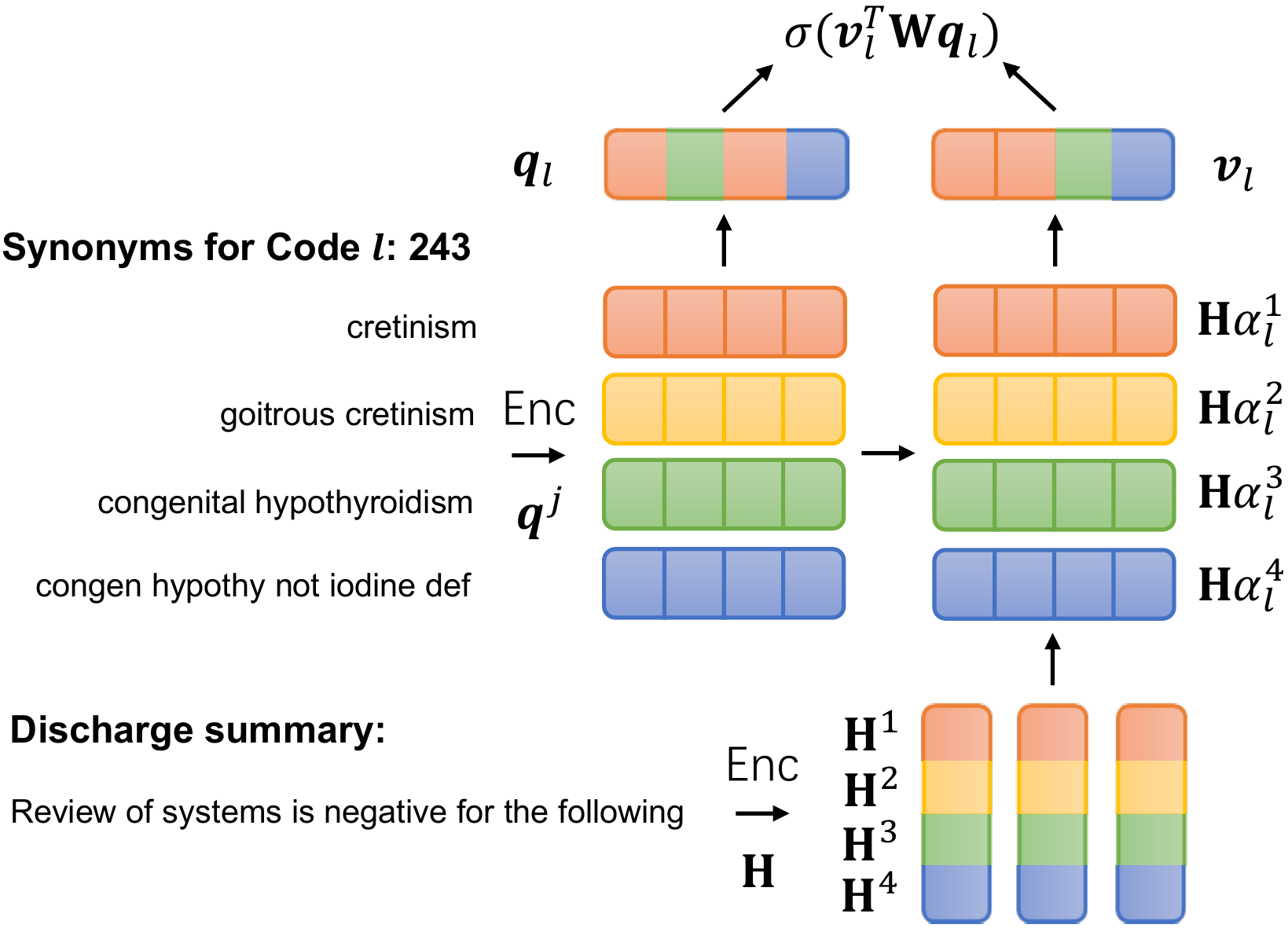}
\caption{The architecture of our proposed MSMN. 
Different colors indicate different code synonyms. We also split hidden representations into different heads for multi-synonyms attention.
}
\label{arch}
\end{figure}

% \subsection{Problem definition}
% Automatic ICD coding can be viewed as a multi-label text classification task.
Consider free text $S$ (usually discharge summaries) from EMR with
% sentences $\{s_i\}_{i=1}^N$, and each sentence consisted of 
words $\{w_{i}\}_{i=1}^{N}$.
% we have multiple code synonyms $\{l^1, l^2, ..., l^m\}$.
The task is to assign a binary label $y_{l} \in \{0,1\}$ based on $S$.
Figure~\ref{arch} shows an overview of our method.

% Our approach apply a shared LSTM \cite{hochreiter1997long} encoder for text encoding and label encoding.
% Multiple label representations are used for 

\subsection{Code Synonyms}

We extend the code description $l^1$ by synonyms from the medical knowledge graph (i.e., UMLS Metathesaurus).
We first align the code to the Concept Unique Identifiers (CUIs) from UMLS.
Then we select corresponding synonyms of English terms from UMLS with the same CUIs and add additional synonyms by removing hyphens and the word ``NOS'' (Not Otherwise Specified).
We denote the code synonyms as $\{l^2, ..., l^M\}$ in which each code synonym $l^j$ is composed of words $\{l_i^j\}_{i=1}^{N_j}$.
% and  are code synonyms from medical knowledge graphs.
% Each code synonym $l^j$ is composed of words $\{l_i^j\}_{i=1}^{N_j}$.

\subsection{Encoding}
% Discharge summaries are much longer than the permitted input length of BERT and 
Previous works \cite{JI2021104998,pascual-etal-2021-towards} have shown that pretrained language models like BERT \cite{devlin-etal-2019-bert} cannot help the ICD coding performance, hence we use an LSTM \cite{hochreiter1997long} as our encoder.
% We use pre-trained word embeddings to map words from texts $w_{i}$ and code synonyms $l^j_i$ to $\mathbf{x}_i, \mathbf{x}^{j}_i$.
We use pre-trained word embeddings to map words $w_{i}$  to $\mathbf{x}_i$.
% where $\mathbf{x}_{ij} \in \mathbb{R}^{l_w}$ is embedding of word $w_{ij}$ and $l_w$ is the word embedding dimension.
A $d$-layer bi-directional LSTM layer takes word embeddings as input to obtain text hidden representations $\mathbf{H} \in \mathbb{R}^h$.
\begin{equation}
    \mathbf{H} = \mathbf{h}_{1}, ..., \mathbf{h}_{N} = {\rm Enc}(\mathbf{x}_{1}, ..., \mathbf{x}_{N})
\end{equation}
For code synonym $l^j$, we apply the same encoder with a max-pooling layer to obtain representation $\mathbf{q}^j \in \mathbb{R}^{h}$.
% \footnote{We have tried using mean pooling or last token representation for $\mathbf{q}^j$ and we observe little performances differences.}:
\begin{equation}
    \mathbf{q}^j = {\rm MaxPool}({\rm Enc}(\mathbf{x}^{j}_{1}, ..., \mathbf{x}^{j}_{N_j}))
\end{equation}

\subsection{Multi-synonyms Attention}
To interact text with multiple synonyms, we propose a multi-synonyms attention inspired by the multi-head attention  \cite{vaswani2017attention}.
We split $\mathbf{H} \in \mathbb{R}^{N \times h}$ into $M$ heads $\mathbf{H}^j \in \mathbb{R}^{N \times \frac{h}{M}}$:
% For code synonym $l^j$, we aggregate the representation $\mathbf{q}^j \in \mathbb{R}^{h}$ using max pooling
% For code $l$, we have $m$ code synonym representations $\mathbf{Q}^l = [\mathbf{q}^1, \mathbf{q}^2, ..., \mathbf{q}^m]$.
% We use multiple label representations as queries and token-level EMRs representations as keys to calculate attention.
\begin{equation}
    \mathbf{H} = \mathbf{H}^1, ..., \mathbf{H}^M
\end{equation}
% Multiple label representations are used as attention queries which are naturally multi-head.
% \begin{equation}
%     \mathbf{A} = {\rm softmax}(\mathbf{Q}{\rm tanh}(\mathbf{W}\mathbf{H}))
% \end{equation}
% Label attention mechanism has been shown successful in ICD coding, since different ICD codes pay attention to different parts of input.
% We consider each ICD code as a query vector $\mathbf{q}_l$, we calculate the attention score between word $w_{ij}$ and ICD code $l$.
% We take each code synonym as an attention query and use 
Then, we use code synonyms $\mathbf{q}^j$ to query $\mathbf{H}^j$. We take the linear transformations 
% ($\mathbf{W}_H$ and $\mathbf{W}_Q$, we omit bias in equation~\ref{eq:sim0}) 
of $\mathbf{H}^j$ and $\mathbf{q}^j$ to calculate attention scores $\mathbf{\alpha}_l^j \in \mathbb{R}^{N}$.
% $\mathbf{\alpha}_l^j$ concentrate on different related text snippets based on different code synonyms.
Text related to code synonym $l^j$ can be represented by $\mathbf{H}\mathbf{\alpha}_l^j$.
We aggregate code-wise text representations $\mathbf{v}_l \in \mathbb{R}^h$ using max-pooling of $\mathbf{H}\mathbf{\alpha}_l^j$ since the text only needs to match one of the synonyms.
% texts related to any code synonym should also relate to the code.
% aggregate from multi-synonyms attention for code-wise text representations $\mathbf{v}_l \in \mathbb{R}^h$.
\begin{align}
    \label{eq:sim0}
    \mathbf{\alpha}_l^j & =  {\rm softmax}(\mathbf{W}_Q\mathbf{q}^j \cdot {\rm tanh}(\mathbf{W}_H\mathbf{H}^j)) \\
    \mathbf{v}_l & = {\rm MaxPool}(\mathbf{H}\mathbf{\alpha}_l^1, ..., \mathbf{H}\mathbf{\alpha}_l^M)
    \label{eq:max}
\end{align}
% Attention score is normalized inside the sentence $s_i$:
% \begin{equation}
%     \alpha_{l,ij} = \frac{\exp(score_{l,ij})}{\sum_k\exp(score_{l,ik})}
% \end{equation}
% Then we can obtain code-aware representations of sentence $s_i$ by weighted summing:
% \begin{equation}
%     \mathbf{v}_{l,i} = \sum_{j}\alpha_{l,ij}\mathbf{h}_{ij}
% \end{equation}
% \begin{equation}
%     \mathbf{v}^l_i = \mathbf{H}\alpha^l_i
% \end{equation}
% \begin{equation}
%     \mathbf{v}_l = \max(\mathbf{v}^l_i)
% \end{equation}

% \subsection{Multiple Descriptions Selection}

\subsection{Classification}
We classify whether the text $S$ contains code $l$ based on the similarity between code-wise text representation $\mathbf{v}_{l}$ and code representation.
We aggregate code synonym representations $\{\mathbf{q}^j\}$ to code representation $\mathbf{q}_l \in \mathbb{R}^h$ by max-pooling.
We then propose using a biaffine transformation to measure the similarity for classification:
% With the code-wise representations $\mathbf{v}_{l}$, we classify label $l$ using similarity via the biaffine transformation 
% % (we omit bias in equation~\ref{eq:biaffine}) 
% of $\mathbf{v}_{l}$ and $\mathbf{q}_l$:
% % On the contrary, we use a biaffine transformation with parameters $\mathbf{W}$ (we omit bias in the equation~\ref{eq:biaffine}) to calculate the similarities.
\begin{align}
    \mathbf{q}_l &= {\rm MaxPool}(\mathbf{q}^1, \mathbf{q}^2, ..., \mathbf{q}^M) \\
    \hat{y}_{l} &= \sigma({\rm logit}_{l}) = \sigma(\mathbf{v}_l^T\mathbf{W}\mathbf{q}_{l})
    \label{eq:biaffine}
\end{align}
Previous works \cite{mullenbach-etal-2018-explainable,ijcai2020-461-vu}  classify codes via\footnote{We omit the biases in all equations for simplification.}:
\begin{equation}
    \hat{y}_{l} = \sigma({\rm logit}_{l}) = \sigma(\mathbf{v}_l^T\mathbf{w}_l)
\end{equation}
Their work need to learn code-dependent parameters $[\mathbf{w}_l]_{l \in \mathcal{C}} \in \mathbb{R}^{\|\mathcal{C}\| \times h}$ for classification, which suffers from training rare codes.
On the contrary, our biaffine function that uses $\mathbf{W}\mathbf{q}_{l}$ instead of $\mathbf{w}_l$ only needs to learn code-independent parameters $\mathbf{W} \in \mathbb{R}^{h \times h}$.

\subsection{Training} % and Inference
We optimize the model using binary cross-entropy between predicted probabilities $\hat{y}_{l}$ and labels $y_l$:
\begin{equation}
    \mathcal{L}
    %_{BCE}(\hat{y}_{l}) 
    = \sum_{l \in \mathcal{C}}-y_l\log(\hat{y}_{l}) - (1-y_l)\log(1-\hat{y}_{l})
\end{equation}

% Since EMRs are too long which have many noisy texts for ICD coding, we further apply R-Drop \cite{liang2021rdrop} to regularize the model to prevent over-fitting by equations~\ref{eq:rdrop}.
% We feed input $S$ to our network twice to obtain two predictions $\{\hat{y}_{l}^0\}$ and $\{\hat{y}_{l}^1\}$ (The differences of two predictions come from different dropouts.).
% The symmetric KL-divergence is used for minimizing differences between two predictions. For the training phase, we jointly minimize the binary cross-entropy losses and the symmetric KL-divergence with a weight $\alpha$.
% \begin{align}
%     \mathcal{L}_{R}(\hat{y}_{l}^0,\hat{y}_{l}^1) = \frac{1}{2}(\mathcal{D}_{KL}(\hat{y}_{l}^0\|\hat{y}_{l}^1) + \mathcal{D}_{KL}(\hat{y}_{l}^1\|\hat{y}_{l}^0)), \\
%     \mathcal{L} = \mathcal{L}_{BCE}(\hat{y}_{l}^0) + \mathcal{L}_{BCE}(\hat{y}_{l}^1) + \alpha \mathcal{L}_{R}(\hat{y}_{l}^0,\hat{y}_{l}^1).
%     \label{eq:rdrop}
% \end{align}
% For the inference phase, we use 0.5 as an unified threshold which means $\hat_{y}_l >= 0.5$

\begin{table}[ht]
    \centering
    \small
    \begin{tabular}{lccc}
    \toprule
    & Train & Dev & Test  \\
    \multicolumn{4}{c}{MIMIC-III Full} \\
    \midrule
    \# Doc. & 47,723 & 1,631 & 3,372 \\
    Avg \# words per Doc. & 1,434 & 1,724 & 1,731 \\  
    Avg \# codes per Doc. & 15.7 & 18.0 & 17.4 \\ 
    Total \# codes & 8,692 & 3,012 & 4,085\\ 
    \midrule
    \multicolumn{4}{c}{MIMIC-III 50} \\
    \midrule
    \# Doc. & 8,066 & 1,573 & 1,729 \\
    Avg \# words per Doc. & 1,478 & 1,739 & 1,763 \\
    Avg \# codes per Doc. & 5.7 & 5.9 & 6.0 \\ 
    Total \# codes & 50 & 50 & 50 \\
    \bottomrule
    
    \end{tabular}
    \caption{Statistics of MIMIC-III dataset under full codes and top-50 codes settings.}
    \label{tab:dataset}
\end{table}

\section{Experiments}

\subsection{Dataset}

MIMIC-III dataset \cite{johnson2016mimic} contains deidentified discharge summaries with human-labeled ICD-9 codes.
We list the document counts, average word counts per document, average codes counts per document, and total codes of the MIMIC-III dataset in Table~\ref{tab:dataset}.
We use the same splits with previous works \cite{mullenbach-etal-2018-explainable,ijcai2020-461-vu}
with two settings as full codes (MIMIC-III full) and top-50 frequent codes (MIMIC-III 50).
We follow the preprocessing of \citet{xie2019ehr} and \citet{ijcai2020-461-vu} to truncate discharge summaries at 4,000 words.
We measure the results using macro AUC, micro AUC, macro $F_1$, micro $F_1$ and precision@k ($k=5$ for MIMIC-III 50, $8$ and $15$ for MIMIC-III full).
% Detailed statistics of the MIMIC-III dataset are listed in Appendix A.

% \subsubsection{Metrics}
% auc f1 p
% To evaluate the explainability of our model, we use two approach to measure the sentence-level predictions.
% (1) Re-train
% (2) Human-label

\subsection{Implementation Details}

We sample $M=4$ and $8$ synonyms per code for MIMIC-III full and MIMIC-III 50 respectively.
We sample synonyms fully randomly from the synonyms set. If some ICD codes do not have enough synonyms, we just repeat these synonyms.
We use the same word embeddings as \citet{ijcai2020-461-vu} which are pretrained on the MIMIC-III discharge summaries using CBOW \cite{mikolov2013efficient} with a hidden size of 100.
We apply R-Drop with $\alpha=5$ \cite{liang2021rdrop} to regularize the model to prevent over-fitting. 
We apply the dropout with a ratio of 0.2 after the word embedding layer and before the classification layer.
For text encoding, we add a linear layer upon the LSTM layer (the output dimension of the linear layer refers to LSTM output dim. in Table~\ref{hyper para}).
We train MSMN with AdamW \cite{loshchilov2017decoupled} with a linear learning rate decay.
We optimize the threshold of classification using the development set.
For the MIMIC-III 50 setting, we train with one 16GB NVIDIA-V100 GPU.
For the MIMIC-III full setting, we train with 8 32GB NVIDIA-V100 GPUs.
We list the detailed training hyper-parameters in Table~\ref{hyper para}.

\begin{table}
\centering
\small
\begin{tabular}{lcc}
\toprule
Parameters & Full & Top 50 \\
\midrule
Emb. dim. & 100 & 100 \\
Emb. dropout & 0.2 & 0.2 \\
LSTM Layer ($d$) & 2 & 1 \\
LSTM hidden dim. & 256 & 512 \\
LSTM output dim. ($h$) & 512 & 512 \\
Synonyms count ($M$) & 4 & 8 \\
Rep. dropout & 0.2 & 0.2 \\
R-Drop weight & 5.0 & 5.0 \\
Epoch & 20 & 20  \\
Peak lr. & 5e-4 & 5e-4 \\
Batch size & 16 & 16 \\
Adam $\epsilon$ & 1e-8 & 1e-8 \\
Weight decay & 0.01 & 0.01 \\
Clipping grad. & 1.0 & 1.0 \\
\bottomrule
\end{tabular}
\caption{Hyper-parameters used for training MIMIC-III full setting and MIMIC-III 50 setting.
}
\label{hyper para}
\end{table}

\subsection{Baselines}
% We list comparable models below.

\noindent\textbf{CAML} \cite{mullenbach-etal-2018-explainable} uses CNN to encode texts and proposes label attention for coding.

\noindent\textbf{MSATT-KG} \cite{xie2019ehr} applies multi-scale attention and GCN to capture codes relations.

\noindent\textbf{MultiResCNN} \cite{li2020icd} encodes text using multi-filter residual CNN.

\noindent\textbf{HyperCore} \cite{cao2020hypercore} embeds ICD codes into the hyperbolic space to utilize code hierarchy and uses GCN to leverage the code co-occurrence.

\noindent\textbf{LAAT} \& \noindent\textbf{JointLAAT} \cite{ijcai2020-461-vu} propose a hierarchical joint learning mechanism to relieve the imbalanced labels, which is our main baseline since it is most similar to our work.

\begin{table*}[h]
    \small 
    \centering
    \begin{tabular}{lcccccc}
    \toprule
    & \multicolumn{2}{c}{AUC}  & \multicolumn{2}{c}{$F_1$} & \multicolumn{2}{c}{Precision@N}\\
    & Macro & Micro & Macro & Micro & P@8 & P@15 \\
    \midrule
    CAML \cite{mullenbach-etal-2018-explainable} & 89.5 & 98.6 & 8.8 & 53.9 &  70.9 & 56.1 \\
    MSATT-KG \cite{xie2019ehr} & 91.0 & \textbf{99.2} & 9.0 & 55.3  & 72.8 & 58.1 \\
    MultiResCNN \cite{li2020icd} & 91.0 & 98.6 & 8.5 & 55.2  & 73.4 & 58.4 \\
    HyperCore \cite{cao2020hypercore} & 93.0 & 98.9 & 9.0 & 55.1  & 72.2 & 57.9 \\
    LAAT \cite{ijcai2020-461-vu} & 91.9 & 98.8 & 9.9 & 57.5  & 73.8 & 59.1 \\
    JointLAAT \cite{ijcai2020-461-vu} & 92.1 & 98.8 & \textbf{10.7} & 57.5  & 73.5 & 59.0  \\
    % PubmedBERT & 87.4 & 98.1 & 4.3 & 44.5 & - & 65.2 & 50.4 \\
    % ISD \cite{zhou2021automatic} & 93.8 & 99.0 & \textbf{11.9} & 55.9  & 74.5 & - \\
    \midrule
    % MSMN \\
    % MultiHead & 94.8 & 99.2 & 9.2 & 57.4 & 81.4 & 74.1 & 59.2 \\
    % MultiLabel & 93.4 & 98.9 & 7.5 & 54.3 & 78.5 & 70.9 & 56.0 \\
    % MultiLabel (F1) & 92.9 & 98.6 & 11.2 & 53.8 & 77.4 & 69.9 & 55.7 \\
    % MSMN & \textbf{94.6} & 99.1 & 9.8 & 56.3 & 73.2 & 58.3 \\
    % MSMN & \textbf{94.4} & 99.1  &  9.4 & 57.2 &  74.4 & 58.9 \\
    % MSMN & \textbf{95.0} & \textbf{99.2}& 7.5& 56.6& \textbf{74.6}& \textbf{59.1} \\
    % MSMN & \textbf{95.1} & \textbf{99.2}& 9.0& \textbf{57.9}& \textbf{74.9}& \textbf{59.6} \\
    % MSMN & \textbf{95.0} & \textbf{99.2}& 9.1& \textbf{57.7}& \textbf{75.3}& \textbf{59.8} \\
    % MSMN & \textbf{95.0} & \textbf{99.2}& 9.7 (10.3)& \textbf{58.0} (58.4, -0.27)& \textbf{75.2}& \textbf{59.9} \\
    MSMN & \textbf{95.0} & \textbf{99.2}& 10.3& \textbf{58.4} & \textbf{75.2}& \textbf{59.9} \\
    % MSMN & \textbf{95.0} & \textbf{99.2} & 9.3 (10.2)& 57.9 (58.3, -0.31) & \textbf{75.3}& \textbf{60.0} \\
    % MSMN & \textbf{94.8} & \textbf{99.1}& 10.5 (10.5) & 57.4 (57.4, 0.00) & \textbf{74.3}& \textbf{59.3} \\
    % 0.9503, 0.9917, 0.0965, 0.5795, 0.8250, 0.7519, 0.5991
    \bottomrule
    \end{tabular}
    \caption{Results on the MIMIC-III full test set.}
    \label{tab:full}
\end{table*}

\begin{table}
    \small 
    \centering
    \begin{tabular}{p{1.58cm}ccccc}
    \toprule
    & \multicolumn{2}{c}{AUC}  & \multicolumn{2}{c}{$F_1$} &\\
     & Macro & Micro & Macro & Micro & P@5 \\
    \midrule
    CAML & 87.5 & 90.9 & 53.2 & 61.4 & 60.9 \\
    MSATT-KG & 91.4 & 93.6 & 63.8 & 68.4 & 64.4  \\
    MultiResCNN & 89.9 & 92.8 & 60.6 & 67.0 & 64.1 \\
    HyperCore & 89.5 & 92.9 & 60.9 & 66.3 & 63.2  \\
    LAAT & 92.5 & 94.6 & 66.6 & 71.5 & 67.5  \\
    JointLAAT & 92.5 & 94.6 & 66.1 & 71.6 & 67.1\\
    % PubmedBERT & 88.6 & 90.8 & 63.3 & 68.1 & 64.4 & - & - \\
    % ISD & \textbf{93.5} & \textbf{94.9} & 67.9 & 71.7 & \textbf{68.2}  \\
    \hline
    % Multi Head & 92.5 & 94.6 & 68.3 & 72.5 & 67.5 & 54.8 & 36.0  \\
    % Multi Label & 92.6 & 94.6 & 67.8 & 71.6 & 67.5 & 54.6 & 35.8 \\
    % Multi Label V2 & 92.7 & 94.7 & 67.9 & 72.0 & 67.9 & 54.8 & 36.1 \\
    % MSMN & 92.8 & 94.7 & 67.8 & \textbf{72.5} & 68.0 \\
    MSMN & \textbf{92.8} & \textbf{94.7} & \textbf{68.3} & \textbf{72.5} & \textbf{68.0} \\
    \bottomrule \\
    \end{tabular}
    \caption{Results on the MIMIC-III 50 test set.}
    \label{tab:50}
\end{table}

\subsection{Main Results}

% \begin{table*}[h]
%     \small 
%     \centering
%     \begin{tabular}{lccccccc}
%     \hline
%     & \multicolumn{2}{c}{AUC}  & \multicolumn{2}{c}{F1} & \multicolumn{3}{c}{Precision@N}\\
%     & Macro & Micro & Macro & Micro & P@5 & P@8 & P@15 \\
%     \hline
%     CAML & 87.5 & 90.9 & 53.2 & 61.4 & 60.9 & - & - \\
%     MSATT-KG & 91.4 & 93.6 & 63.8 & 68.4 & 64.4 & - & - \\
%     MultiResCNN & 89.9 & 92.8 & 60.6 & 67.0 & 64.1 & - & - \\
%     HyperCore & 89.5 & 92.9 & 60.9 & 66.3 & 63.2 & - & - \\
%     LAAT & 92.5 & 94.6 & 66.6 & 71.5 & 67.5 & \textbf{54.7} & \textbf{35.7} \\
%     JointLAAT & 92.5 & 94.6 & 66.1 & 71.6 & 67.1 & 54.6 & \textbf{35.7} \\
%     % PubmedBERT & 88.6 & 90.8 & 63.3 & 68.1 & 64.4 & - & - \\
%     ISD & \textbf{93.5} & \textbf{94.9} & \textbf{67.9} & \textbf{71.7} & \textbf{68.2} & - & - \\
%     \hline
%     % Multi Head & 92.5 & 94.6 & 68.3 & 72.5 & 67.5 & 54.8 & 36.0  \\
%     % Multi Label & 92.6 & 94.6 & 67.8 & 71.6 & 67.5 & 54.6 & 35.8 \\
%     % Multi Label V2 & 92.7 & 94.7 & 67.9 & 72.0 & 67.9 & 54.8 & 36.1 \\
%     MSMN & 92.8 & 94.7 & 67.8 & 72.5 & 68.0 & 54.8 & 36.0 \\
%     \hline \\
%     \end{tabular}
%     \caption{Test set results for MIMIC-III 50 datasets.
% }
%     \label{tab:50}
% \end{table*}

Table~\ref{tab:full} and \ref{tab:50} show the main results under the MIMIC-III full and MIMIC-III 50 settings, respectively. 
% Introduction of compared methods are in Appendix C.
Under the full setting, our MSMN achieves 95.0 (+2.0), 99.2 (+0.0), 10.3 (-0.4), 58.4 (+0.9), 75.2 (+1.4), and 59.9 (+0.8) in terms of macro-AUC, micro-AUC, macro-$F_1$, micro-$F_1$, P@8, and P@15 respectively (parentheses shows the differences against previous best results), which shows that MSMN obtains state-of-the-art results in most metrics.
Under the top-50 codes setting, MSMN performs better than LAAT  in all metrics and achieves state-of-the-art scores of 92.8 (+0.3), 94.7 (+0.1), 68.3 (+1.7), 72.5 (+0.9), 68.0 (+0.5) on macro-AUC, micro-AUC, macro-$F_1$, micro-$F_1$, and P@5, respectively. 
We notice that the macro $F_1$ has a large variance in every epoch under the MIMIC-III full setting since it is more sensitive in a long tail problem.

% l and $F_1$ can have different orders among results which is mainly because $F_1$ is threshold-related.

\subsection{Discussion}

To explore the influence of leveraging different numbers of code synonyms, we search $M$ among $\{1,2,4,8,16\}$ on the MIMIC-III 50 dataset. Results are shown in Table \ref{tab:ab}.
Compared with $M=1$ that we only use the original ICD code descriptions, leveraging more synonyms from UMLS consistently improves the performance. Using $M=4,8$ achieves the best performance in terms of AUC, and $M=8$ achieves the best performance in terms of $F_1$ and P@5. In addition, the median and mean count of UMLS synonyms are 5.0 and 5.4 respectively, which echoes why the results of $M=4$ or $8$ are better.
% Noticing MSMN perform better in top-50 codes 
% Top-50 codes have more synonyms than rare codes which can be benefited from MSMN much.

% Therefore, considering the number of codes to be predicted, we choose $m=8$ for top-50 codes setting and choose $m=4$ for full codes setting.

% \noindent\textbf{R-Drop} We remove R-Drop and find the model performs worse in all metrics which implies that it is necessary to regularize MSMN due to the noisy discharge summaries. We select the default $\alpha=5$ from R-Drop without a grid search.

To evaluate the effectiveness of our proposed biaffine-based similarity function, we compare it with the baseline LAAT in Table \ref{tab:ab}. We also provide a simple function by removing $\mathbf{W}$ to $\mathbf{v}_l^T\mathbf{q}_{l}$ in Equation~\ref{eq:biaffine}. Results show that the biaffine-based similarity scoring performs best among others.

% \noindent\textbf{Similarity Scoring} We replace the similarity function by removing $\mathbf{W}$ to $\mathbf{v}_l^T\mathbf{q}_{l}$ or using LAAT style \cite{ijcai2020-461-vu} ($\mathbf{v}_l^T\mathbf{w}_{l}$).

\begin{figure}[t]
\centering
\includegraphics[width=2.6in]{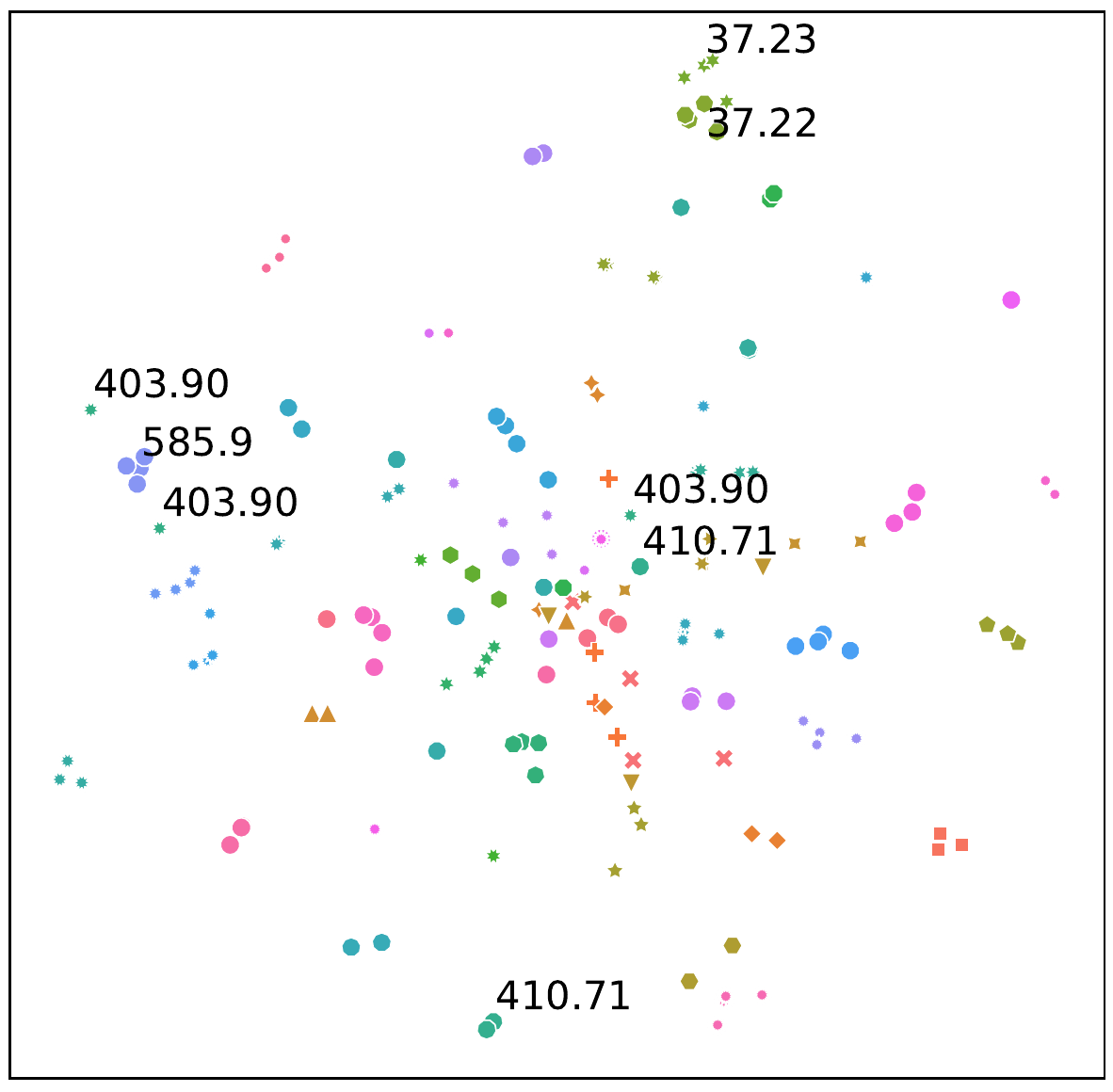}
\caption{T-SNE visualization of code synonym representations learned from MIMIC-III 50.}
\label{fig:example}
\end{figure}

To better understand what MSMN learns from the multi-synonyms attention, we plot the synonym representations $\mathbf{q}^j$ under MIMIC-III 50 setting via t-SNE \cite{JMLR:v9:vandermaaten08a} in Figure~\ref{fig:example}.
We observe for some codes like \textit{585.9} (``chronic kidney diseases''), all synonym representations cluster together, which indicates that synonyms extract similar text snippets.
However, codes like \textit{410.71} (``subendocardial infarction initial episode of care'' or ``subendo infarct, initial'') and \textit{403.90} (``hypertensive chronic kidney disease, unspecified, with chronic kidney disease stage i through stage iv'' or ``unspecified orhy kid w cr kid i iv'') with very different synonyms learn different representations, which benefits to match different text snippets.
Furthermore, we observe it has similar representations for sibling codes \textit{37.22} (``left heart cardiac catheterization'') and \textit{37.23} (``rt/left heart card cath''), which indicates the model can also implicitly capture the code hierarchy.

\begin{table}
    \small 
    \centering
    \begin{tabular}{lccccc}
    \toprule
    & \multicolumn{2}{c}{AUC}  & \multicolumn{2}{c}{$F_1$} & \\
    & Macro & Micro & Macro & Micro & P@5 \\
    \midrule
    % $m=1$ & 92.1 & 94.2 & 65.5 & 70.4 & 67.0 \\
    $M=1$ & 92.1 & 94.2 & 67.4 & 71.0 & 67.0 \\
    % $m=2$ & 92.6 & 94.6 & 67.5 & 71.6 & 67.2 \\
    $M=2$ & 92.6 & 94.6 & 67.6 & 71.7 & 67.2 \\
    % $m=4$ & \textbf{92.8} & \textbf{94.7} & 68.0 & 71.8 & 67.7 \\
    $M=4$ & \textbf{92.8} & \textbf{94.7} & 67.9 & 71.9 & 67.7 \\
    $\underline{M=8}$ & \textbf{92.8} & \textbf{94.7} & \textbf{68.3} & \textbf{72.5} & \textbf{68.0} \\
    % $m=16$ & 92.5 & 94.6 & 67.4 & 71.6 & 67.6 \\
    $M=16$ & 92.5 & 94.6 & 66.9 & 71.5 & 67.6 \\
    % \hline
    % $\underline{\alpha=5}$ & \textbf{92.8} & \textbf{94.7} & \textbf{68.3} & \textbf{72.5} & \textbf{68.0} \\
    % $\alpha=0$ & 92.5 & 94.4 & 68.0 & 71.8 & 67.1 \\
    % -est cls & \\
    \midrule
    $\underline{\mathbf{v}_l^T\mathbf{W}\mathbf{q}_{l}}$ & \textbf{92.8} & \textbf{94.7} & \textbf{68.3} & \textbf{72.5} & \textbf{68.0} \\
    % $\mathbf{v}_l^T\mathbf{q}_{l}$ & 92.4& 94.4& 67.1& 70.9& 66.5 \\
    % $\mathbf{v}_l^T\mathbf{w}_{l}$ & 91.5& 94.1& 63.9& 70.2& 66.1 \\
    $\mathbf{v}_l^T\mathbf{q}_{l}$ & 92.5& 94.5& 67.1& 71.2& 67.1 \\
    $\mathbf{v}_l^T\mathbf{w}_{l}$ & 91.5& 94.1& 65.1& 70.8& 66.3 \\
    \bottomrule
    \end{tabular}
    \caption{Results of different settings including synonyms counts and scoring functions on MIMIC-III 50 dataset.
    Underlined setting denotes the default parameters used in MSMN.}
    \label{tab:ab}
\end{table}

\subsection{Memory Complexity}
The memory usage of our proposed MSMN is dominated by Equation~\ref{eq:sim0} and Equation~\ref{eq:max}.
We suppose batch size as $B$, word count as $N$, label count as $C$ and synonyms count as $M$.
Calculating Equation~\ref{eq:sim0} for all $j$ simultaneously requires calculating Einstein summation \cite{daniel2018opt} among tensors with shape $B \times N \times h$ and shape $C \times M \times h$ to shape $B \times C \times N \times M$.
Calculating Equation~\ref{eq:max} requires calculating Einstein summation among tensors with shape $B \times N \times h$ and shape $B \times C \times N \times M$ to shape $B \times C \times h \times M$. The memory complexities of these two equations are linearly proportional to $M$.

\section{Related Work}
Automatic ICD coding is an important task in the medical NLP community.
Earlier works use machine learning methods for coding \cite{larkey1996combining,pestian-etal-2007-shared,perotte2014diagnosis}.
With the development of neural networks, many recent works consider ICD coding as a multi-label text classification task. They usually apply RNN or CNN to encode texts and use the label attention mechanism to extract and match the most relevant parts for classification.
The label attention relies on the label representations as attention queries.
\citet{li2020icd,ijcai2020-461-vu} randomly initialize the label representations which ignore the code semantic information.
\citet{cao2020hypercore} use the average of word embeddings as label representations to leverage the code semantic information.
\citet{xie2019ehr,cao2020hypercore} use GCN to fuse hierarchical structures of ICD codes for label representations.
% \citet{sonabend2020automated} propose unsupervised xxx.
\citet{kim2021read} use CNN for codes representations.
Compared with previous works, 
% our method uses a shared encoder for texts and codes which implies texts and codes should have similar representations.
% Furthermore, 
we use synonyms instead of a single description to represent the code, which can provide more comprehensive expressions of codes.

Biomedical entity linking is a related task to automatic ICD coding. The task requires standardizing given terms to a pre-defined concept dictionary. There are two differences between biomedical entity linking and automatic ICD coding: (1) They have different target concepts. ICD coding map EMRs to ICD codes, while biomedical entity linking usually map terms to a larger dictionary like SNOMED-CT or UMLS. (2) They have different input formats. Entity linking task has labeled entities in texts, while ICD coding only provides texts. Synonyms have also been used in biomedical entity linking \cite{sung-etal-2020-biomedical,yuan2021coder}.
BioSYN \cite{sung-etal-2020-biomedical} uses marginalization to sum the probabilities of all synonyms as the similarity between a term and a concept.
However, we consider multi-synonyms attention to extracting different parts of clinical texts to interact with synonyms.

\section{Conclusions}
In this paper, we propose MSMN to leverage code synonyms from UMLS to improve the automatic ICD coding.
Multi-synonyms attention is proposed for extracting different related text snippets for code-wise text representations.
We also propose a biaffine transformation to calculate similarities among texts and codes for classification.
Experiments show that MSMN outperforms previous methods with label attention and achieves state-of-the-art results in the MIMIC-III dataset.
Ablation studies show the effectiveness of multi-synonyms attention and biaffine-based similarity.
% Despite the training samples from MIMIC-III, code synonyms help coders to know how diseases are described in EMRs which we prove can also be helpful in automatic ICD coding.
% Our proposed MSMN shows the abilities of code synonyms in automatic ICD coding.
% We believe code synonyms should be further explored for better few or zero-shot ICD coding.
% We obtain code synonyms from UMLS to extend the ICD code descriptions.

\section*{Acknowledgements}
We would like to thank the anonymous reviewers for their helpful comments and suggestions.
We thank Fuli Luo, Shengxuan Luo, Hongyi Yuan, Xu Chen, and Jiayu Li for their help.
This work was supported by Alibaba Group through Alibaba Research Intern Program.

% \section*{Ethical Considerations}
% This work 

\bibliographystyle{acl_natbib}
\bibliography{anthology,acl2021}

\begin{thebibliography}{31}
\expandafter\ifx\csname natexlab\endcsname\relax\def\natexlab#1{#1}\fi

\bibitem[{Bodenreider(2004)}]{Bodenreider2004}
Olivier Bodenreider. 2004.
\newblock The unified medical language system (umls): integrating biomedical
  terminology.
\newblock \emph{Nucleic acids research}, 32(suppl\_1):D267--D270.

\bibitem[{Cao et~al.(2020)Cao, Chen, Liu, Zhao, Liu, and
  Chong}]{cao2020hypercore}
Pengfei Cao, Yubo Chen, Kang Liu, Jun Zhao, Shengping Liu, and Weifeng Chong.
  2020.
\newblock Hypercore: Hyperbolic and co-graph representation for automatic icd
  coding.
\newblock In \emph{Proceedings of the 58th Annual Meeting of the Association
  for Computational Linguistics}, pages 3105--3114.

\bibitem[{Daniel et~al.(2018)Daniel, Gray et~al.}]{daniel2018opt}
G~Daniel, Johnnie Gray, et~al. 2018.
\newblock Opt$\backslash$\_einsum-a python package for optimizing contraction
  order for einsum-like expressions.
\newblock \emph{Journal of Open Source Software}, 3(26):753.

\bibitem[{de~Lima et~al.(1998)de~Lima, Laender, and
  Ribeiro-Neto}]{de1998hierarchical}
Luciano~RS de~Lima, Alberto~HF Laender, and Berthier~A Ribeiro-Neto. 1998.
\newblock A hierarchical approach to the automatic categorization of medical
  documents.
\newblock In \emph{Proceedings of the seventh international conference on
  Information and knowledge management}, pages 132--139.

\bibitem[{Devlin et~al.(2019)Devlin, Chang, Lee, and
  Toutanova}]{devlin-etal-2019-bert}
Jacob Devlin, Ming-Wei Chang, Kenton Lee, and Kristina Toutanova. 2019.
\newblock \href {https://doi.org/10.18653/v1/N19-1423} {{BERT}: Pre-training of
  deep bidirectional transformers for language understanding}.
\newblock In \emph{Proceedings of the 2019 Conference of the North {A}merican
  Chapter of the Association for Computational Linguistics: Human Language
  Technologies, Volume 1 (Long and Short Papers)}, pages 4171--4186,
  Minneapolis, Minnesota. Association for Computational Linguistics.

\bibitem[{Falis et~al.(2019)Falis, Pajak, Lisowska, Schrempf, Deckers, Mikhael,
  Tsaftaris, and O’Neil}]{falis2019ontological}
Matus Falis, Maciej Pajak, Aneta Lisowska, Patrick Schrempf, Lucas Deckers,
  Shadia Mikhael, Sotirios Tsaftaris, and Alison O’Neil. 2019.
\newblock Ontological attention ensembles for capturing semantic concepts in
  icd code prediction from clinical text.
\newblock In \emph{Proceedings of the Tenth International Workshop on Health
  Text Mining and Information Analysis (LOUHI 2019)}, pages 168--177.

\bibitem[{Hochreiter and Schmidhuber(1997)}]{hochreiter1997long}
Sepp Hochreiter and J{\"u}rgen Schmidhuber. 1997.
\newblock Long short-term memory.
\newblock \emph{Neural computation}, 9(8):1735--1780.

\bibitem[{Ji et~al.(2021)Ji, Hölttä, and Marttinen}]{JI2021104998}
Shaoxiong Ji, Matti Hölttä, and Pekka Marttinen. 2021.
\newblock \href
  {https://doi.org/https://doi.org/10.1016/j.compbiomed.2021.104998} {Does the
  magic of bert apply to medical code assignment? a quantitative study}.
\newblock \emph{Computers in Biology and Medicine}, 139:104998.

\bibitem[{Johnson et~al.(2016)Johnson, Pollard, Shen, Li-Wei, Feng, Ghassemi,
  Moody, Szolovits, Celi, and Mark}]{johnson2016mimic}
Alistair~EW Johnson, Tom~J Pollard, Lu~Shen, H~Lehman Li-Wei, Mengling Feng,
  Mohammad Ghassemi, Benjamin Moody, Peter Szolovits, Leo~Anthony Celi, and
  Roger~G Mark. 2016.
\newblock Mimic-iii, a freely accessible critical care database.
\newblock \emph{Scientific data}, 3(1):1--9.

\bibitem[{Kim and Ganapathi(2021)}]{kim2021read}
Byung-Hak Kim and Varun Ganapathi. 2021.
\newblock Read, attend, and code: Pushing the limits of medical codes
  prediction from clinical notes by machines.
\newblock In \emph{Machine Learning for Healthcare Conference}, pages 196--208.
  PMLR.

\bibitem[{Larkey and Croft(1996)}]{larkey1996combining}
Leah~S Larkey and W~Bruce Croft. 1996.
\newblock Combining classifiers in text categorization.
\newblock In \emph{Proceedings of the 19th annual international ACM SIGIR
  conference on Research and development in information retrieval}, pages
  289--297.

\bibitem[{Li and Yu(2020)}]{li2020icd}
Fei Li and Hong Yu. 2020.
\newblock Icd coding from clinical text using multi-filter residual
  convolutional neural network.
\newblock In \emph{Proceedings of the AAAI Conference on Artificial
  Intelligence}, volume~34, pages 8180--8187.

\bibitem[{Liang et~al.(2021)Liang, Wu, Li, Wang, Meng, Qin, Chen, Zhang, and
  Liu}]{liang2021rdrop}
Xiaobo Liang, Lijun Wu, Juntao Li, Yue Wang, Qi~Meng, Tao Qin, Wei Chen, Min
  Zhang, and Tie-Yan Liu. 2021.
\newblock R-drop: Regularized dropout for neural networks.
\newblock In \emph{NeurIPS}.

\bibitem[{Loshchilov and Hutter(2019)}]{loshchilov2017decoupled}
Ilya Loshchilov and Frank Hutter. 2019.
\newblock Decoupled weight decay regularization.
\newblock In \emph{7th International Conference on Learning Representations,
  {ICLR} 2019, New Orleans, LA, USA, May 6-9, 2019}.

\bibitem[{Mikolov et~al.(2013)Mikolov, Chen, Corrado, and
  Dean}]{mikolov2013efficient}
Tom{\'{a}}s Mikolov, Kai Chen, Greg Corrado, and Jeffrey Dean. 2013.
\newblock \href {http://arxiv.org/abs/1301.3781} {Efficient estimation of word
  representations in vector space}.
\newblock In \emph{1st International Conference on Learning Representations,
  {ICLR} 2013, Scottsdale, Arizona, USA, May 2-4, 2013, Workshop Track
  Proceedings}.

\bibitem[{Mullenbach et~al.(2018)Mullenbach, Wiegreffe, Duke, Sun, and
  Eisenstein}]{mullenbach-etal-2018-explainable}
James Mullenbach, Sarah Wiegreffe, Jon Duke, Jimeng Sun, and Jacob Eisenstein.
  2018.
\newblock \href {https://doi.org/10.18653/v1/N18-1100} {Explainable prediction
  of medical codes from clinical text}.
\newblock In \emph{Proceedings of the 2018 Conference of the North {A}merican
  Chapter of the Association for Computational Linguistics: Human Language
  Technologies, Volume 1 (Long Papers)}, pages 1101--1111, New Orleans,
  Louisiana. Association for Computational Linguistics.

\bibitem[{Pascual et~al.(2021)Pascual, Luck, and
  Wattenhofer}]{pascual-etal-2021-towards}
Damian Pascual, Sandro Luck, and Roger Wattenhofer. 2021.
\newblock \href {https://doi.org/10.18653/v1/2021.bionlp-1.6} {Towards
  {BERT}-based automatic {ICD} coding: Limitations and opportunities}.
\newblock In \emph{Proceedings of the 20th Workshop on Biomedical Language
  Processing}, pages 54--63, Online. Association for Computational Linguistics.

\bibitem[{Perotte et~al.(2014)Perotte, Pivovarov, Natarajan, Weiskopf, Wood,
  and Elhadad}]{perotte2014diagnosis}
Adler Perotte, Rimma Pivovarov, Karthik Natarajan, Nicole Weiskopf, Frank Wood,
  and No{\'e}mie Elhadad. 2014.
\newblock Diagnosis code assignment: models and evaluation metrics.
\newblock \emph{Journal of the American Medical Informatics Association},
  21(2):231--237.

\bibitem[{Pestian et~al.(2007)Pestian, Brew, Matykiewicz, Hovermale, Johnson,
  Cohen, and Duch}]{pestian-etal-2007-shared}
John~P. Pestian, Chris Brew, Pawel Matykiewicz, DJ~Hovermale, Neil Johnson,
  K.~Bretonnel Cohen, and Wlodzislaw Duch. 2007.
\newblock \href {https://www.aclweb.org/anthology/W07-1013} {A shared task
  involving multi-label classification of clinical free text}.
\newblock In \emph{Biological, translational, and clinical language
  processing}, pages 97--104, Prague, Czech Republic. Association for
  Computational Linguistics.

\bibitem[{Sonabend et~al.(2020)Sonabend, Cai, Ahuja, Ananthakrishnan, Xia, Yu,
  and Hong}]{sonabend2020automated}
Aaron Sonabend, Winston Cai, Yuri Ahuja, Ashwin Ananthakrishnan, Zongqi Xia,
  Sheng Yu, and Chuan Hong. 2020.
\newblock Automated icd coding via unsupervised knowledge integration (unite).
\newblock \emph{International journal of medical informatics}, 139:104135.

\bibitem[{Song et~al.(2020)Song, Zhang, Sadoughi, Xie, and
  Xing}]{ijcai2020-556}
Congzheng Song, Shanghang Zhang, Najmeh Sadoughi, Pengtao Xie, and Eric Xing.
  2020.
\newblock \href {https://doi.org/10.24963/ijcai.2020/556} {Generalized
  zero-shot text classification for icd coding}.
\newblock In \emph{Proceedings of the Twenty-Ninth International Joint
  Conference on Artificial Intelligence, {IJCAI-20}}, pages 4018--4024.
  International Joint Conferences on Artificial Intelligence Organization.
\newblock Main track.

\bibitem[{Sung et~al.(2020)Sung, Jeon, Lee, and
  Kang}]{sung-etal-2020-biomedical}
Mujeen Sung, Hwisang Jeon, Jinhyuk Lee, and Jaewoo Kang. 2020.
\newblock \href {https://doi.org/10.18653/v1/2020.acl-main.335} {Biomedical
  entity representations with synonym marginalization}.
\newblock In \emph{Proceedings of the 58th Annual Meeting of the Association
  for Computational Linguistics}, pages 3641--3650, Online. Association for
  Computational Linguistics.

\bibitem[{Suo et~al.(2018)Suo, Ma, Yuan, Huai, Zhong, Gao, and
  Zhang}]{suo2018deep}
Qiuling Suo, Fenglong Ma, Ye~Yuan, Mengdi Huai, Weida Zhong, Jing Gao, and
  Aidong Zhang. 2018.
\newblock Deep patient similarity learning for personalized healthcare.
\newblock \emph{IEEE transactions on nanobioscience}, 17(3):219--227.

\bibitem[{Sutton et~al.(2020)Sutton, Pincock, Baumgart, Sadowski, Fedorak, and
  Kroeker}]{sutton2020overview}
Reed~T Sutton, David Pincock, Daniel~C Baumgart, Daniel~C Sadowski, Richard~N
  Fedorak, and Karen~I Kroeker. 2020.
\newblock An overview of clinical decision support systems: benefits, risks,
  and strategies for success.
\newblock \emph{NPJ digital medicine}, 3(1):1--10.

\bibitem[{van~der Maaten and Hinton(2008)}]{JMLR:v9:vandermaaten08a}
Laurens van~der Maaten and Geoffrey Hinton. 2008.
\newblock \href {http://jmlr.org/papers/v9/vandermaaten08a.html} {Visualizing
  data using t-sne}.
\newblock \emph{Journal of Machine Learning Research}, 9(86):2579--2605.

\bibitem[{Vaswani et~al.(2017)Vaswani, Shazeer, Parmar, Uszkoreit, Jones,
  Gomez, Kaiser, and Polosukhin}]{vaswani2017attention}
Ashish Vaswani, Noam Shazeer, Niki Parmar, Jakob Uszkoreit, Llion Jones,
  Aidan~N Gomez, {\L}ukasz Kaiser, and Illia Polosukhin. 2017.
\newblock Attention is all you need.
\newblock In \emph{Advances in neural information processing systems}, pages
  5998--6008.

\bibitem[{Vu et~al.(2020)Vu, Nguyen, and Nguyen}]{ijcai2020-461-vu}
Thanh Vu, Dat~Quoc Nguyen, and Anthony Nguyen. 2020.
\newblock \href {https://doi.org/10.24963/ijcai.2020/461} {A label attention
  model for icd coding from clinical text}.
\newblock In \emph{Proceedings of the Twenty-Ninth International Joint
  Conference on Artificial Intelligence, {IJCAI-20}}, pages 3335--3341.
\newblock Main track.

\bibitem[{Xie and Xing(2018)}]{xie2018neural}
Pengtao Xie and Eric Xing. 2018.
\newblock A neural architecture for automated icd coding.
\newblock In \emph{Proceedings of the 56th Annual Meeting of the Association
  for Computational Linguistics (Volume 1: Long Papers)}, pages 1066--1076.

\bibitem[{Xie et~al.(2019)Xie, Xiong, Yu, and Zhu}]{xie2019ehr}
Xiancheng Xie, Yun Xiong, Philip~S Yu, and Yangyong Zhu. 2019.
\newblock Ehr coding with multi-scale feature attention and structured
  knowledge graph propagation.
\newblock In \emph{Proceedings of the 28th ACM International Conference on
  Information and Knowledge Management}, pages 649--658.

\bibitem[{Yuan et~al.(2022)Yuan, Zhao, Sun, Li, Wang, and Yu}]{yuan2021coder}
Zheng Yuan, Zhengyun Zhao, Haixia Sun, Jiao Li, Fei Wang, and Sheng Yu. 2022.
\newblock \href {https://doi.org/https://doi.org/10.1016/j.jbi.2021.103983}
  {Coder: Knowledge-infused cross-lingual medical term embedding for term
  normalization}.
\newblock \emph{Journal of Biomedical Informatics}, page 103983.

\bibitem[{Zhou et~al.(2021)Zhou, Cao, Chen, Liu, Zhao, Niu, Chong, and
  Liu}]{zhou2021automatic}
Tong Zhou, Pengfei Cao, Yubo Chen, Kang Liu, Jun Zhao, Kun Niu, Weifeng Chong,
  and Shengping Liu. 2021.
\newblock Automatic icd coding via interactive shared representation networks
  with self-distillation mechanism.
\newblock In \emph{Proceedings of the 59th Annual Meeting of the Association
  for Computational Linguistics and the 11th International Joint Conference on
  Natural Language Processing (Volume 1: Long Papers)}, pages 5948--5957.

\end{thebibliography}

% \appendix
% \section{MIMIC-III Dataset Statistics}
% \label{appendix:stat}

% \section{Training Details}
% \label{appendix:para}

% \noindent\textbf{ISD} \cite{zhou2021automatic} uses shared representations to help long-tail codes, and learns from code descriptions using self-distillation.

% \noindent\textbf{Fine-tune Language Model} \cite{ji2021does} fine-tunes PubmedBERT \cite{gu2020domain} for ICD coding, they deal with long document input by chunking.

\end{document}